\documentclass[conference]{IEEEtran}
\IEEEoverridecommandlockouts

\usepackage{comment}
\usepackage{algorithm}
\usepackage{multirow}
\usepackage{algpseudocode}
\usepackage{booktabs}
\usepackage{url}
\usepackage{cite}
\usepackage{amsmath,amssymb,amsfonts}
\usepackage{graphicx}
\usepackage{textcomp}
\usepackage{xcolor}
\def\BibTeX{{\rm B\kern-.05em{\sc i\kern-.025em b}\kern-.08em
    T\kern-.1667em\lower.7ex\hbox{E}\kern-.125emX}}
\begin{document}

\title{Model-free Speculative Decoding for Transformer-based ASR with Token Map Drafting\\
}

\author{
\IEEEauthorblockN{Tuan Vu Ho, Hiroaki Kokubo, Masaaki Yamamoto, and Yohei Kawaguchi}
\IEEEauthorblockA{\textit{Hitachi, Ltd.}, Tokyo, Japan \\
tuanvu.ho.zt@hitachi.com, hiroaki.kokubo.dz@hitachi.com, \\masaaki.yamamoto.af@hitachi.com, yohei.kawaguchi.xk@hitachi.com}
}
\maketitle

\begin{abstract}
End-to-end automatic speech recognition (ASR) systems based on transformer architectures, such as Whisper, offer high transcription accuracy and robustness. 
However, their autoregressive decoding is computationally expensive, hence limiting deployment on CPU-based and resource-constrained devices. 
Speculative decoding (SD) mitigates this issue by using a smaller draft model to propose candidate tokens, which are then verified by the main model. However, this approach is impractical for devices lacking hardware accelerators like GPUs.
To address this, we propose \emph{Token Map Drafting}, a model-free SD technique that eliminates the need for a separate draft model. 
Instead, we leverage a precomputed n-gram token map derived from domain-specific training data, enabling efficient speculative decoding with minimal overhead. 
Our method significantly accelerates ASR inference in structured, low-perplexity domains without sacrificing transcription accuracy.
Experimental results demonstrate decoding speed-ups of $1.27\times$ on the CI-AVSR dataset and $1.37\times$ on our internal dataset without degrading recognition accuracy.
Additionally, our approach achieves a $10\%$ absolute improvement in decoding speed over the Distill-spec baseline running on CPU, highlighting its effectiveness for on-device ASR applications.
\end{abstract}


\section{Introduction}
The adoption of encoder-decoder transformers \cite{vaswani2017attention} has significantly advanced automatic speech recognition (ASR), yielding substantial improvements in transcription accuracy \cite{radford2023robust, gulati2020conformer, ao2022speecht5}. These models utilize autoregressive decoding to systematically generate tokens, leveraging strong contextual awareness to enhance recognition performance. However, autoregressive inference is computationally expensive, particularly in resource-constrained environments, such as mobile devices that require real-time processing.

To address this challenge, various optimization techniques have been proposed to accelerate transformer inference. These include efficient attention mechanisms \cite{dao2022flashattention, dao2023flashattention2, li2024folding}, model quantization \cite{dettmers2022gpt3, Rybakov20232bitCQ}, and knowledge distillation \cite{wang2020minilm, jiao2019tinybert}, each aiming to reduce computational overhead while maintaining transcription accuracy.

Speculative decoding (SD) \cite{leviathan2023fast} is a simple technique aims at accelerating autoregressive transformer inference by leveraging parallelized token proposals. 
It employs lightweight draft models to predict speculative token sequences, which are subsequently verified and corrected by the primary model. 
This approach has been further explored and enhanced in subsequent research. For instance, \cite{chen2023accelerating} proposed speculative sampling methods to expedite large language model decoding, while \cite{Liu2024online} focused on continuously updating draft models on observed user queries, reducing distribution shifts and enhancing prediction accuracy.

Recent works, such as \cite{gandhi2023distil} and \cite{segal2024whisper}, have explored SD in ASR by incorporating mechanisms like knowledge distillation and multi-head verification. 
While effective in large-scale systems, running an additional draft model locally increases computational overhead, potentially introducing more latency instead of improving decoding speed.
The reliance on fine-tuned draft models therefore limits the scalability of these methods to lightweight, CPU-based deployments.

Drawing from statistical language modeling, we propose \textit{Token Map Drafting}, a model-free SD method that leverages precomputed draft tokens derived from domain-specific transcriptions.
Our approach generates draft tokens via an efficient n-gram token map, eliminating the need for an additional draft model.
As such, it is particularly suited for structured ASR tasks in resource-constrained environments. 
Furthermore, structured text contexts characterized by low perplexity, such as maintenance commands or repetitive sequences, enhance its effectiveness.

In summary, our key contributions are: (1) we propose the \textit{Token Map Drafting} model to enhance the inference speed of Whisper model on resource-constrained device, and (2) we conducted analysis to achieve optimal speculative decoding speed-up on domain-specific dataset.

\section{Speculative Decoding}
\begin{figure}[bt]
    \centering
    \includegraphics[width=\linewidth]{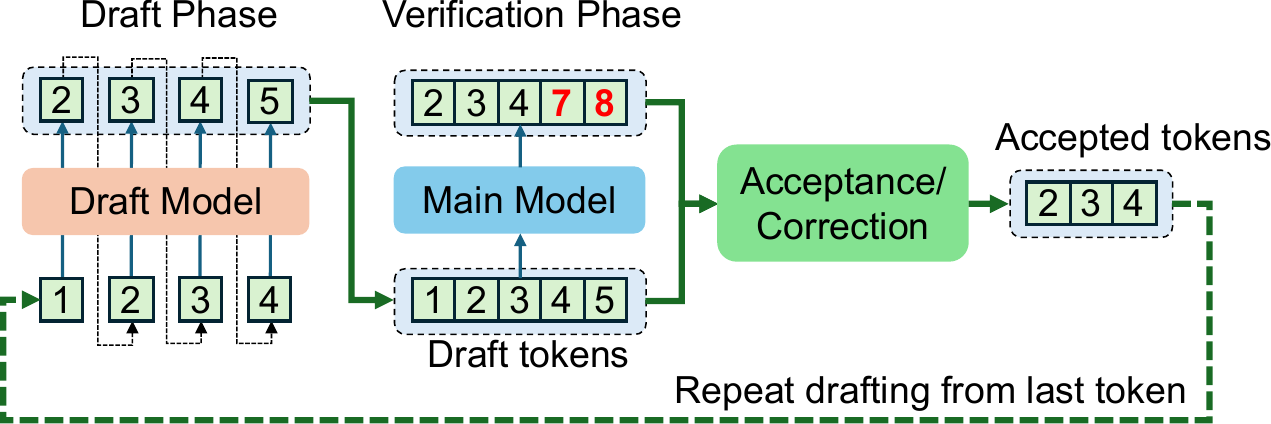}
    \caption{Overview of speculative decoding. Divergent tokens is highlight in red, which is removed after Acceptance/Correction step.}
    \label{fig:speculative_decoding_baseline}
\end{figure}

Introduced by \cite{leviathan2023fast}, SD is a technique designed to accelerate autoregressive language models by utilizing a lightweight draft decoder to generate multiple candidate tokens in parallel.
In conventional autoregressive inference, generating $K$ tokens requires $K$ sequential runs of the model.
SD leverages the observation that language modeling tasks often contain simpler subtasks that can be well-approximated by a smaller model. Thus, a draft token sequence is first generated autoregressively using a smaller model and then verified in parallel by the main model in a single step. 
Since most of the draft token sequence is expected to match the final output of the main model, SD can significantly reduce inference time compared to conventional autoregressive decoding.
Figure \ref{fig:speculative_decoding_baseline} illustrates the processing steps of speculative decoding. 

The detailed steps of speculative decoding are as follows:
\begin{enumerate}
    \item \textbf{Draft model prediction:} A lightweight draft model (e.g. distilled model) generates a speculative sequence of $k$ tokens given the current context $x$. This step is computationally inexpensive and helps in reducing the overall decoding time:
    \begin{equation}
        \hat{y}_{1:k} = \text{DraftModel}(x)
    \end{equation}
    The draft model predicts tokens in parallel based on statistical patterns or simplified learned representations.
    
    \item \textbf{Verification by the main model:} The primary (larger) model evaluates the draft sequence token by token by computing the conditional probability of each predicted token:
    \begin{equation}
        P(y_i | x, y_{1:i-1}) \quad \forall i \in \{1, ..., k\}
    \end{equation}
    This step ensures that the speculative tokens comply with the main model expectations.
    If a token probability is above a predefined threshold, it is considered correct and accepted. 
    Otherwise, the decoding process returns to the predictions of the main model itself.
    
    \item \textbf{Acceptance and correction:} If the verification step confirms the draft tokens, they are accepted, and the decoding moves to the next segment. However, if any token does not meet the required confidence threshold, the main model generates a replacement token. This correction step ensures that errors from the draft model do not propagate.
    
    \item \textbf{Continuation until completion:} Steps 1–3 are repeated iteratively until the full sequence is generated or an end-of-sequence (EOS) token is reached. If the speculative model provides high-quality predictions, the overall decoding process speeds up significantly while maintaining accuracy.
    
    \item \textbf{Finalize output:} The final output sequence is constructed from the combination of accepted speculative tokens and any necessary corrections from the main model. The resulting sequence is computationally efficient and aligned with the quality expected from the primary model.
\end{enumerate}

In the field of ASR, speculative decoding was explored in \cite{gandhi2023distil} through knowledge distillation for the Whisper model, demonstrating substantial decoding speed-ups. 
In this approach, a lightweight distilled model, such as distill-large-v3, serves as the draft model for the main model (e.g., Whisper-large-v3). 
Candidate tokens are generated using standard autoregressive decoding by the draft model.
Since the distilled model and the main model share the same encoder structure, the verification step is performed only on the decoder of the main model.  

Similarly, \cite{segal2024whisper} extended speculative decoding by introducing a multi-head decoder with an additional $K$ heads, enabling the prediction of $K+1$ tokens per step to improve efficiency. The $K+1$ draft tokens are then verified using the base head, similar to conventional speculative decoding.  
Trained on LibriSpeech \cite{panayotov2015librispeech} and VoxPopuli \cite{wang2021voxpopuli}, these models achieve a 50\% reduction in latency while maintaining competitive accuracy.  

\section{Proposed methodology}
\subsection{Speculative decoding with \textit{Token Map Drafting}}
\begin{figure*}
    \centering
    \includegraphics[width=0.7\linewidth]{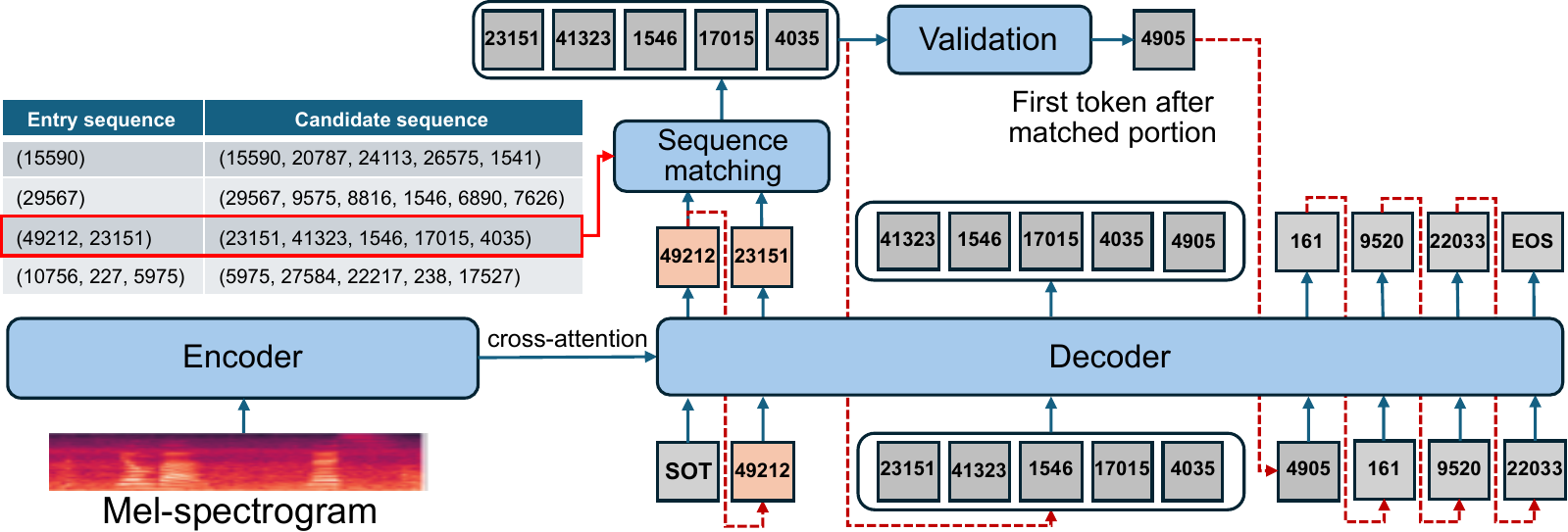}
    \caption{Overview of our proposed speculative decoding with \textit{Token Map Drafting}. 'SOT' represents the Start Of Transcript token, and 'EOS' denotes the end-of-sequence token.}
    \label{fig:proposed_speculative_decoding}
\end{figure*}
While SD effectively accelerates inference, conventional approaches depend on fine-tuning a draft model or modifying the model architecture, making them impractical for resource-constrained devices such as mobile platforms. However, many ASR applications in industrial domains operate within constrained vocabularies and structured sentence patterns, such as in-car commands or measurement descriptions. In such cases, the draft sequence can often be inferred from the first few generated tokens.

We propose a model-free SD method that leverages a precomputed \textit{token map} to efficiently generate draft sequences without requiring an additional draft model
By analyzing domain-specific text corpora, we construct a dictionary where each key represents an n-gram token sequence, and the corresponding value contains high-confidence candidate sequences that frequently follow it.
Instead of sequentially generating tokens via a draft model, our approach selects speculative sequences from this token map, reducing computational overhead while maintaining decoding accuracy. 
This method is particularly effective in structured ASR tasks, such as Whisper-based transcription of repetitive commands.

Figure~\ref{fig:proposed_speculative_decoding} demonstrate an example of our proposed SD method.
The encoder processes a Mel-spectrogram input, and the decoder employs cross-attention. 
The decoding process begins with an entry sequence and a candidate sequence of tokens.
Both sequences are expanded speculatively, and a sequence matching process identifies shared segments between them. 
Following validation, the tokens after the matched portion are determined, and the decoding process continues iteratively.

\subsection{Token map construction}
To build the token map, we begin by converting each sentence into a sequence of tokens using a pretrained tokenizer.
The token map functions as a dictionary containing key-value pairs. 
Each key represents an n-gram sequence, and the corresponding value is a list of token sequences that follow the key sequence in the dataset.
The length of n-gram sequence can vary 1 to $N$.

\begin{figure}[bt]
    \centering
    \includegraphics[width=0.8\linewidth]{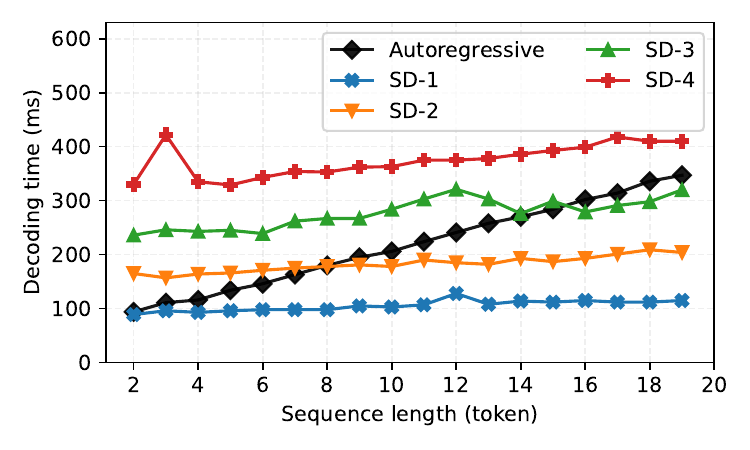}
    \caption{Decoding time comparison between standard autoregressive decoding and SD. SD-$K$ represents speculative decoding with $K$ candidate token sequences.}
    \label{fig:decoding_time_batch_vs_length}
\end{figure}

To optimize decoding efficiency, we rank and prune candidate token sequences based on token length and frequency thresholds, prioritizing longer sequences to maximize speedup while omitting low-confidence candidates. 
Since SD does not always outperform standard autoregressive decoding in every scenario, we empirically analyze inference time under varying candidate numbers and sequence lengths. 
Figure~\ref{fig:decoding_time_batch_vs_length} shows that SD achieves speedup only when candidate sequence lengths exceed certain thresholds (e.g., 9 tokens for 2 candidates, 16 for 3 candidates). 
Beyond 3 candidates, SD show less effective. 
Based on these findings, we iteratively merge candidates with their nearest matches until the condition is met.
When the decoder output misaligns with a speculative candidate, we handle mismatches by truncating the decoder state at the first unmatched token and resuming AR decoding until the next n-gram match is found.
This ensures minimal additional delay without requiring a full re-run of the decoder.

\begin{figure}[bt]
    \centering
    \includegraphics[width=0.9\linewidth]{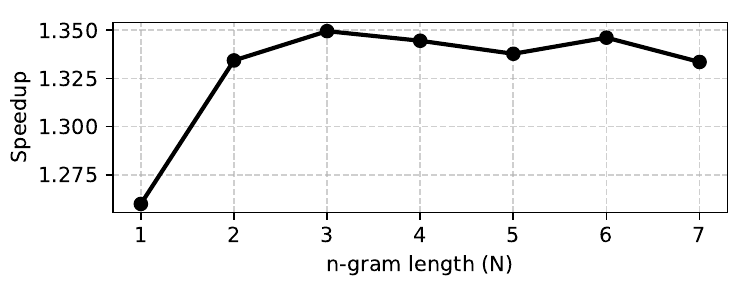}
    \caption{Speedup analysis of processing times with varying n-gram length $N$.}
    \label{fig:Speedup_vs_ngram}
\end{figure}

To determine the optimal n-gram length $N$, we measure speedup rates across different values of $N$ using Whisper-small on an internal dataset.
Figure~\ref{fig:Speedup_vs_ngram} shows that speedup peaks at $N=3$, which we adopt for subsequent experiments.

We summarize our proposed method in the below pseudo-algorithm:
\begin{algorithm}
\caption{Token Map Construction Algorithm}
\begin{algorithmic}[1]
\Require Training transcriptions $\mathcal{D}$
\Ensure Token map $\mathcal{T}$
\For {sentence $s \in \mathcal{D}$}
    \State Tokenize $s$ using pre-trained tokenizer
    \State Extract n-grams ($n=1,2,...,N$) from tokens
    \State Add extracted n-grams to token tree $\mathcal{T}$ as key

\EndFor
\For{n-gram $k$ from token map $\mathcal{T}$ key}
    \State Extract all token sequences in $\mathcal{D}$ that follow $k$ as candidates sequence
    \State Add extracted candidates sequences to $\mathcal{T}[key]$
\EndFor
\State Prune $\mathcal{T}$ based on condition in Figure \ref{fig:decoding_time_batch_vs_length}
\State \Return $\mathcal{T}$

\end{algorithmic}
\end{algorithm}

\begin{table*}[t]
\centering
\caption{Comparison of different draft models in speculative decoding. The table highlights differences in speedup ($S$), acceptance rate ($A_r$), and average acceptance length ($A_l$) across datasets.}
\label{tab:draft_model_comparison}

\begin{tabular}{@{}ccccccccc@{}}
\toprule
\multirow{3}{*}{Method} & \multirow{3}{*}{Main Model} & \multirow{3}{*}{Draft Model} & \multicolumn{3}{c}{CI-AVSR Dataset} & \multicolumn{3}{c}{Internal Dataset} \\
\cmidrule(lr){4-6} \cmidrule(lr){7-9}
                        &                             &                              & $S$ & $A_r$ & $A_l$ & $S$ & $A_r$ & $A_l$ \\
\midrule
Baseline                & Whisper-large-v3           & Whisper-turbo-v3            & 1.02 & 44.1 & 3.46 & 1.27 & 85.8 & 8.41 \\
\midrule
\multirow{3}{*}{Proposed method} 
                        & Whisper-large-v3           & Token map                   & \textbf{1.27} & 38.6 & 3.42 & \textbf{1.37} & 85.6 & \textbf{3.92} \\
                        & Whisper-medium             & Token map                   & 1.12 & \textbf{56.0} & 2.32 & 1.36 & 95.9 & 3.95 \\
                        & Whisper-small              & Token map                   & 1.09 & 51.1 & \textbf{1.83} & 1.35 & \textbf{97.1} & 3.96 \\
\bottomrule
\end{tabular}%

\end{table*}

\section{Experiments}
In this section, we describe our experimental setup for evaluating the performance of our proposed method against the baseline. All experiments were conducted on a laptop with Intel Core i5 CPU. The implementation was developed in $C++$ using the CTranslate2 \footnote{\url{https://github.com/OpenNMT/CTranslate2/}} library.

\subsection{Datasets and Setup}
We evaluate the performance of speculative decoding using two datasets:

\begin{itemize}
    \item \textbf{CI-AVSR:}\cite{Dai2022CIAVSRAC} A dataset designed for continuous integration of audio-visual speech recognition, containing diverse utterances across various speakers and environments.
    \item \textbf{Our Internal Dataset:} A specialized dataset consisting of phrases from maintenance tasks. Each phrase typically includes a device or part name followed by measurement values at the end of the phrase, simulating real-world structured speech input.
\end{itemize}

As a baseline method, we compare with the Distill-spec method \cite{gandhi2023distil} using the Whisper-large-v3 \footnote{https://huggingface.co/openai/whisper-large-v3} along with Whisper-large-v3-turbo \footnote{https://huggingface.co/openai/whisper-large-v3-turbo} as a draft model, as Distil-Whisper-large-v3 only supports the English language. 
For our proposed method, we use different Whisper variants, including Whisper-large-v3, Whisper-medium, and Whisper-small to evaluate the performance.

\subsection{Evaluation metrics}
To assess the effectiveness of speculative decoding, we use the following benchmark metrics:

\begin{itemize}
    \item \textbf{Speedup ($S$):} Measures the relative improvement in decoding time compared to baseline autoregressive decoding.
    \begin{equation}
        S = \frac{T_{\text{baseline}}}{T_{\text{speculative}}}
    \end{equation}
    where $T_{\text{baseline}}$ is the decoding time using standard autoregressive inference and $T_{\text{speculative}}$ is the decoding time using speculative decoding.
    
    \item \textbf{Acceptance rate ($A_r$):} The proportion of speculative tokens accepted by the main model:
    \begin{equation}
        A_r = \frac{N_{\text{accepted}}}{N_{\text{total}}}
    \end{equation}
    where $N_{\text{accepted}}$ is the number of speculative tokens accepted, and $N_{\text{total}}$ is the total number of speculative tokens proposed.
    
    \item \textbf{Average acceptance length ($A_l$):} The average number of consecutive speculative tokens accepted before a correction is required:
    \begin{equation}
        A_l = \frac{\sum L_{\text{accepted}}}{N_{\text{sequences}}}
    \end{equation}
    where $L_{\text{accepted}}$ is the number of speculative tokens accepted in a sequence, and $N_{\text{sequences}}$ is the total number of decoded sequences.
\end{itemize}

\subsection{Results}
Table~\ref{tab:draft_model_comparison} compares the performance of different draft models in speculative decoding, highlighting speedup ($S$), acceptance rate ($A_r$), and average acceptance length ($A_l$) across two datasets. 
The proposed token map method consistently outperforms the Whisper-turbo draft model in terms of speedup, particularly on the Internal Dataset, where it achieves up to 1.37$\times$ speedup compared to 1.27$\times$ for the baseline. 
This suggests that the token map method is more effective at accelerating inference in structured domain-specific transcriptions.

Despite its efficiency, the token map method exhibits a lower acceptance rate ($A_r$) on the CI-AVSR dataset, which may indicate challenges in adapting to more diverse or spontaneous speech patterns. 
The lower average acceptance length ($A_l$) compared to the Whisper-turbo model on the Internal Dataset suggests that speculative token predictions are more conservative, possibly reducing the risk of incorrect acceptance. 
Overall, these results demonstrate that our proposed model-free approach achieves competitive or superior performance without requiring an additional draft model, making it well-suited for deployment on resource-constrained devices.

\section{Conclusion}
This paper introduced Token Map Drafting, a model-free speculative decoding technique designed to accelerate autoregressive ASR inference in structured, low-perplexity domains. By utilizing a precomputed n-gram token map instead of a separate draft model, our approach significantly reduces computational overhead, achieving up to 1.37× speedup on a domain-specific dataset and demonstrating a 10\% absolute improvement over the distill-spec baseline on CPU-based systems. These results highlight the method's potential for real-time, on-device ASR applications, particularly in resource-constrained environments.

While our experimental results suggest that the proposed method maintains transcription accuracy, we recognize that the accuracy of the main model plays a critical role in decoding efficiency, particularly under conditions where the main model may introduce recognition errors which cause inconsistency with the Token Map. Future work will investigate the relationship between main model accuracy and decoding performance to better understand these dynamics and further optimize our approach.

\bibliographystyle{IEEEtran}
\bibliography{mybib}

\end{document}